\documentclass[10pt,twocolumn,letterpaper]{article}

\usepackage{cvpr}
\usepackage{times}
\usepackage{epsfig}
\usepackage{graphicx}
\usepackage{amsmath}
\usepackage{amssymb}
\usepackage{multirow}

\graphicspath{ {./img/} }

\usepackage[breaklinks=true,bookmarks=false]{hyperref}

\cvprfinalcopy 

\def\cvprPaperID{****} 

\begin{document}
\hyphenation{ColorUNet}
\title{\textit{ColorUNet}: A convolutional classification approach to colorization \\[.2in] \normalsize CS231n Final Project, Stanford University}

\author{Vincent Billaut\\
Department of Statistics\\
Stanford University\\
{\tt\small vbillaut@stanford.edu}
\and
Matthieu de Rochemonteix\\
Department of Statistics\\
Stanford University\\
{\tt\small mderoche@stanford.edu}
\and
Marc Thibault\\
ICME\\
Stanford University\\
{\tt\small marcthib@stanford.edu}
}

\maketitle



\begin{abstract}
This paper tackles the challenge of colorizing grayscale images. We take a deep convolutional neural network approach, and choose to take the angle of classification, working on a finite set of possible colors. Similarly to a recent paper, we implement a loss and a prediction function that favor realistic, colorful images rather than ``true'' ones.

We show that a rather lightweight architecture inspired by the U-Net, and trained on a reasonable amount of pictures of landscapes, achieves satisfactory results on this specific subset of pictures. We show that data augmentation significantly improves the performance and robustness of the model, and provide visual analysis of the prediction confidence.

We show an application of our model, extending the task to video colorization. We suggest a way to smooth color predictions across frames, without the need to train a recurrent network designed for sequential inputs.

\end{abstract}

\section*{Introduction}

The problem of colorization is one that comes quickly to mind when thinking about interesting challenges involving pictural data. Namely, the goal is to build a model that takes as an input the greyscale version of an image (or even an actual ``black and white'' picture) and outputs its colorized version, as close to the original as possible (or at least realistic, if the original is not in colors). This problem is complex and interesting for several reasons, as the final output needs to be an image of the same dimension as the input image. We want to train a model that is able to recognize shapes that are typical of a category of items and apply the appropriate colorization.

One clear upside to this challenge is that any computer vision dataset, and even any image bank really, is a proper dataset for the colorization problem (the image itself is the model's expected output, and its greyscale version is its input).
Given a grayscale image, our algorithm outputs the same image, colorized. A conversion of the images to the YUV format allows an easy formulation of the problem in terms of a reconstitution of the U and V channels\footnote{The transformation between RGB and YUV color encodings is a linear transformation. In the YUV system, Y is the luminance (grayscale) and U,V encode the colors.}.

We formulate the colorization problem as a classification problem to gain flexibility in the prediction process and output more colorful images. We aim at reproducing state of the art results that give vivid, visually appealing results, with a much smaller network.

One of our main concerns is to design our own architecture from scratch, rather than using parts from other architecture or pre-trained portions. This is to ensure modularity, full control of the design and to keep the overall complexity reasonable.

\section{Related Work} \label{relatedwork}

Historically, older approaches of the colorization problem use an additional input, a \textit{seed scribble} from the user to propagate colors, as does \cite{levin2004colorization}. It may also be seen as a corollary of color style transfer algorithms using a similar image as a ``seed'' as in \cite{he2017neuralct}. Both approaches yield very good results but need supervision from the user for each image. In this project, we are interested in fully automated (re)colorization.

Classical approaches to this task, \eg \cite{cheng2015deep} and \cite{dahl2016tinyclouds}, aim at predicting an image as close as possible to the ground truth, and notably make use of a simple $L_2$ loss in the YUV space, which penalizes predictions that fall overall too far from the ground truth. As a consequence, the models trained following such methods usually tend to be very conservative and to give desaturated, pale results. Usually, the images need some postprocessing adjustments as in \cite{deshpande2015learning} to have a realistic aspect.

On the contrary, authors of \cite{zhang2016colorful} take another angle and set their objective to be ``\textit{plausible} colorization'' (and not necessarily \textit{accurate}), which they validate with a large-scale human trial. To achieve such results, they formulate the colorization task as a classification task using color bins, as suggested already in \cite{charpiat2008automatic}.

Their approach is the most appealing as they have colorful results and we found the classification formulation to be interesting. However, the network architecture they use is heavy -- constraining to small batch sizes to fir into a single GPU --, and the scale of their training set -- several millions of images -- is prohibitive. The reason behind this complexity is that in order to encode meaningful features that help to colorize the image, the receptive field has to be large. The approach of the article is to downsample the image a lot in the intermediate layers and then upsample it using Transpose Convolution layers. To keep a lot of information in the intermediate layers, the number of filters in the model of \cite{zhang2016colorful} has to be significant, resulting in a large, slow and expensive training.

Authors of \cite{ronneberger2015unet} have shown that connections between hidden layers of a bottleneck neural network could enhance the performance greatly, by injecting locational information in the upsampling process, and improving the gradient flow.
We hope that applying this method will allow us to train a colorizing model more quickly and more efficiently, with less parameters, and on a smaller dataset.

Part of the challenges that are interesting but not yet tackled in the literature involve videos. General information propagation frameworks in a video involving bilateral networks as discussed in \cite{jampani2017video} could be seen as a good starting point to implement consistent colorization of picture sequences, if we manage to embed the colorizing information and then propagate it as any other information. The work realized in \cite{zhu2017video} is also interesting since it tackles video stylization by grouping the frames, choosing a representative frame for each group and using the output of the network on that frame as a guideline, which enhances temporal consistency greatly. However, the adaptation of such an algorithm to the much more complex task of image colorization is far beyond the scope of this project.

Actually, one promising way to perform image colorization is to be able to learn meaningful color-related representations for the images (which often involves using very deep and heavy or pretrained architecture as in \cite{larsson2016repres}) and then ensure the temporal consistency of them.

Given our commitment to developing a lightweight model, we prefered focusing on an efficient colorization for images and then add a layer of temporal convolution to stabilize the video colorization.

\section{Methods} \label{methods}

\subsection{Colorization as classification}
\begin{figure}
\begin{center}
\includegraphics[width=200px]{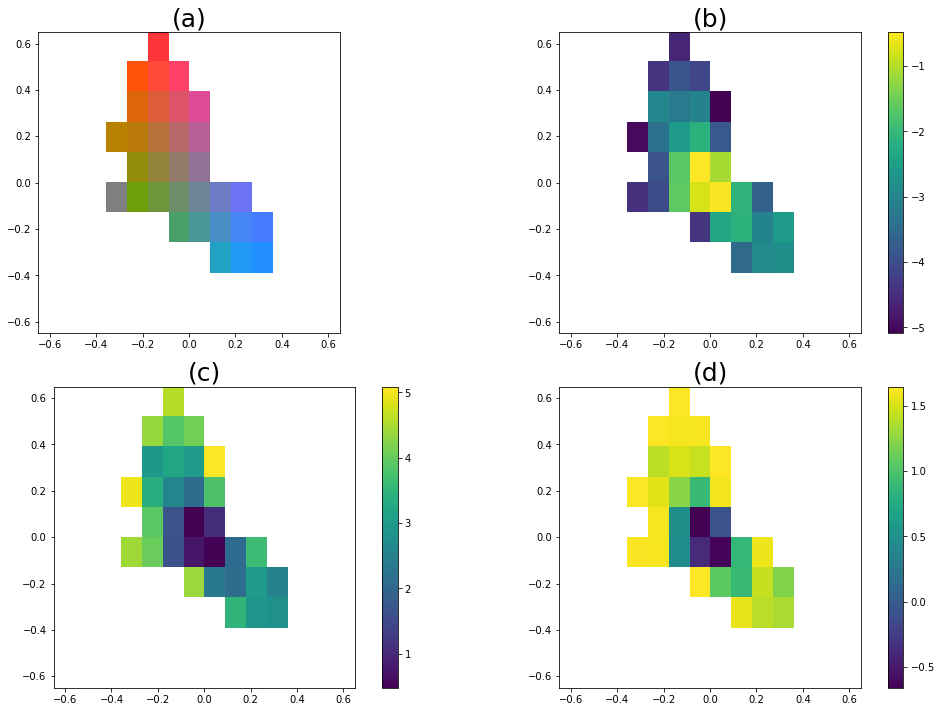}
\caption{(a) Color map, showing the mean color (\textit{chrominance}) of each of the selected bins (here, we set a threshold of 32 bins to select). (b) Frequency map (log-scale), shows the empirical frequency of the colors within each bin, computed over a dataset beforehand. (c) Inverse-frequency map (log-scale), \ie the inverse of (b). (d) Weight map (log-scale), shows the weights assigned to each bin after rebalancing. Interestingly, we notice that the amplitude in weights is much smaller than the amplitude in frequency (2 orders of magnitude against 4), which means that we partially make up for the underrepresentation bias and will therefore encourage the prediction of rare colors.}
\label{cdex}
\end{center}
\end{figure}
\begin{figure}
\begin{center}
\includegraphics[width=200px]{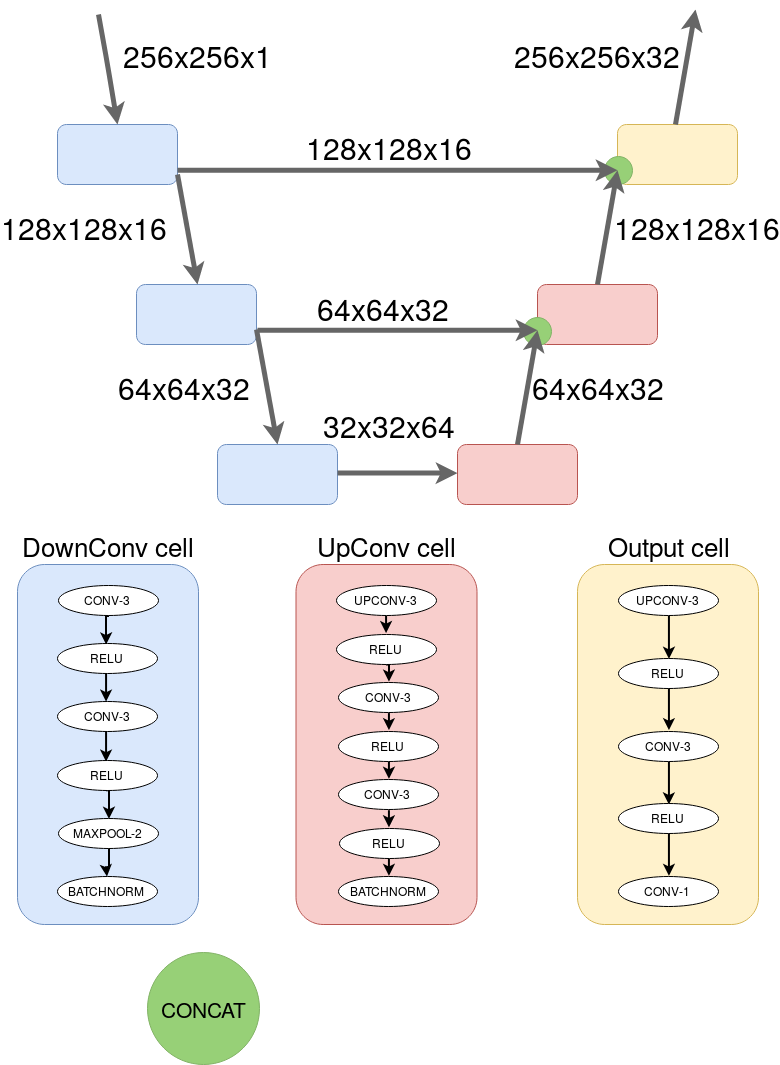}
\caption{Structure of the ColorUNet. We use 3 types of cells: DownConv Cells that use 2 stacked convolutional layers to have a large perceptive field and a maxpooling to downsample the image, UpConv cells that use 1 ConvTranspose Layer to upsample the image and then 2 convolutional layers, and an Output cell that is a simplified version of the UpConv cell. All the convolutional layers have kernel size 3$\times$3  with stride 1 and padding 1. The ConvTranspose layers have kernel size 3$\times$3 with stride 2 and padding 1. The maxpooling layers have size 2$\times$2 and stride 2, with no padding. }
\label{structure}
\end{center}
\end{figure}
As we discussed in section \ref{relatedwork}, we are taking the angle of \cite{zhang2016colorful}. As a result, we are approaching the colorization as a classification problem, and therefore using a (weighted) cross-entropy loss.

Concretely, we want to discretize our colorspace, and for that we simply split our colormap into equal sized bins. As a first step, in order to reduce the computational toll, we don't want too many bins and are therefore restricting our discretization to the $n$ most frequent color bins, as learned on a large dataset beforehand. In what follows, if a color from our actual image does not fall within one of our $n$ bins, we will simply assign it to the closest available. This simplification leads to a rather faint degradation of the images.

Following the approach of \cite{zhang2016colorful}, we want to boost the possibility of a rare color being predicted, and therefore reproduce the following tricks:
\begin{itemize}
\item Use \textit{rebalancing} to give larger weights to rare colors in our loss function. Precisely, each pixel value $p$, assigned to its closest bin $b \in \mathbb{R}^n$ is given a weight $w_p$ such that $$w_p \propto \left( (1-\lambda) \widehat{P}(b) + {\lambda \over n} \right)^{-1}$$ where $\widehat{P}(b)$ is the estimated probability of the bin (computed prior to our model's training) and $\lambda \in [0,1]$ a tuning parameter (the closer to 1, the less we take the bin's frequency into account).
\item Use an \textit{annealed-mean} to output predictions $y$ from the probability distribution $\textbf{Z}$ over our $n$ bins to the full original color space. The idea is to find a compromise between taking the color class with the maximum probability (the mode), which gives a rather colorful result but sometimes lacking spatial consistency, and taking the weighted average over the bins, which gives a rather flat, sepia kind of tone. To achieve this we use a temperature parameter $T > 0$ in the following softmax-like formula for one pixel $$y = f_T(\textbf{z}) = {\exp(\log(\textbf{z}) / T) \over \sum_i \exp(\log(\textbf{z}_i) / T)}$$ where $\textbf{z}$ is the $n$-dimensional probability vector of a given pixel over the $n$ bins, and the sum in the denominator is over all the bins.
\end{itemize}

Figure \ref{cdex} shows our discretized colorspace, as well as what the weights of the bins look like after rebalancing.
\begin{figure}
\begin{center}
\includegraphics[width=200px]{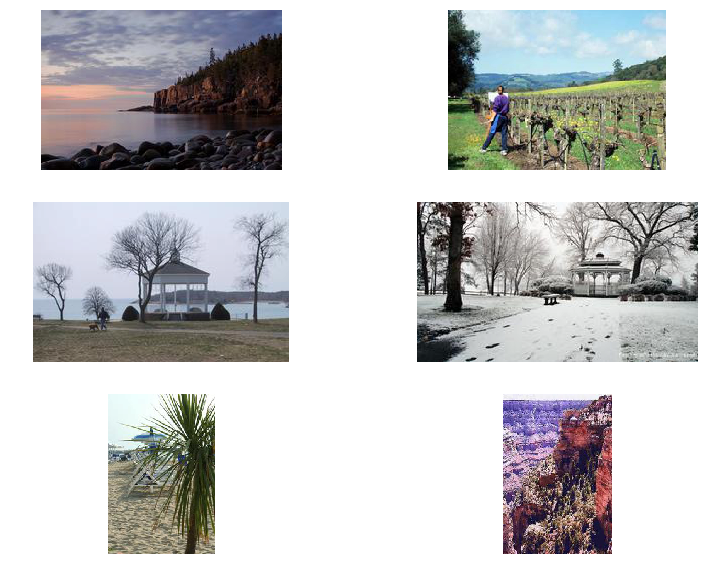}
\caption{Sample images from our dataset (here, the validation set). We included categories such as \texttt{coast}, \texttt{beach} and \texttt{gazebo\_exterior} from SUN, and categories like \texttt{landscape}, \texttt{field} and \texttt{canyon} from ImageNet.}
\label{sampletrain}
\end{center}
\end{figure}

\subsection{Neural architecture: ColorUNet}

Our task has now become a segmentation problem, for which we have to predict the correct class (representing a color bin) out of 32, for every pixel in the input image. Our final model is a U-Net that has 3 downsampling groups of convolutional layers and 2 connections between hidden layers of the same size (we set aside the last and first layers of the downsampling process).

Figure \ref{structure} summarizes the structure of the final network, that we refer to as ColorUNet.

To end up with this final model, we have tried several architectures. We started by trying a simple ``flat'' convolutional network with 3 layers. However, this model gave very poor performance. Intuitively, what happens is that the spatial receptive field was not large enough to capture meaningful general shapes.
\begin{figure*}
\begin{center}
\includegraphics[width=450px]{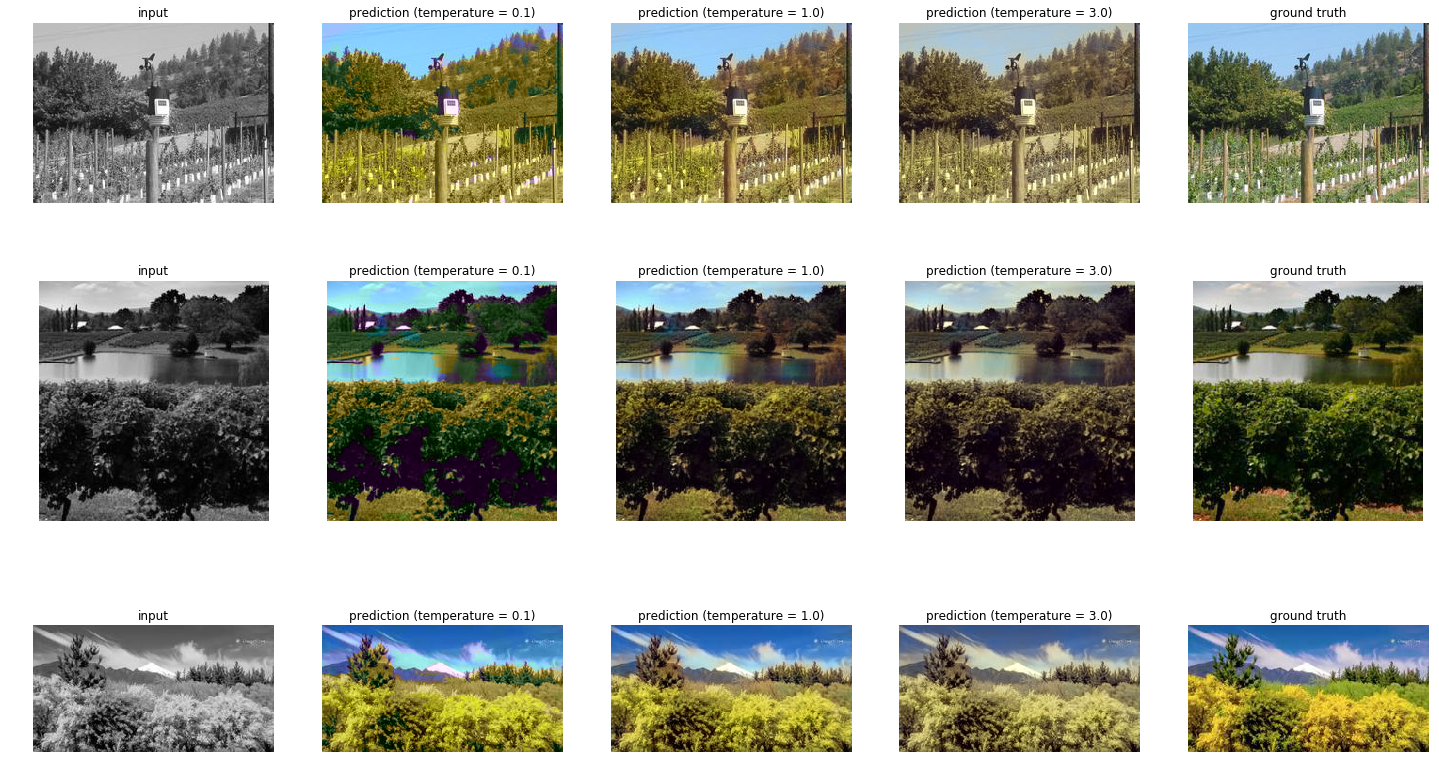}
\caption{Sample predictions of the ColorUNet on the validation set, for several prediction temperatures. The temperature parameter allows a tradeoff between vivid colors and elimination of artifacts of a wrong color.}
\label{good}
\end{center}
\end{figure*}
Building upon this idea, we then implemented a second model, that is deeper, with 6 groups of convolutional layers. The structure is a \textit{bottleneck}, where 3 groups of layers downsample the image using a max pooling, and 3 groups of layers upsample the image. Using groups of 3$\times$3 convolutional layers rather than bigger filters allows to have a large receptive field with fewer parameters.The \textit{bottleneck} structure is a way to constrain the model to embed recognizible object shapes and the associated color.

This network gave interesting results but it was very unstable and slow to train, which is easily explained by the depth of the architecture. The quality of the output we obtained was not satisfying -- even for a reduced task --, because of this underfitting. Furthermore, it would have needed more filters in the hidden layers to tackle the upsampling part of the flow, as we need to encode more information in the smaller layers to correctly upsample the image afterwards.

Those observations led us to design our final ColorUNet, that still has a bottleneck structure but uses connections across layers to improve the gradient flow and help the upsampling. We also added additional batch normalization layers to improve training stability.

\begin{figure*}
\begin{center}
\includegraphics[width=450px]{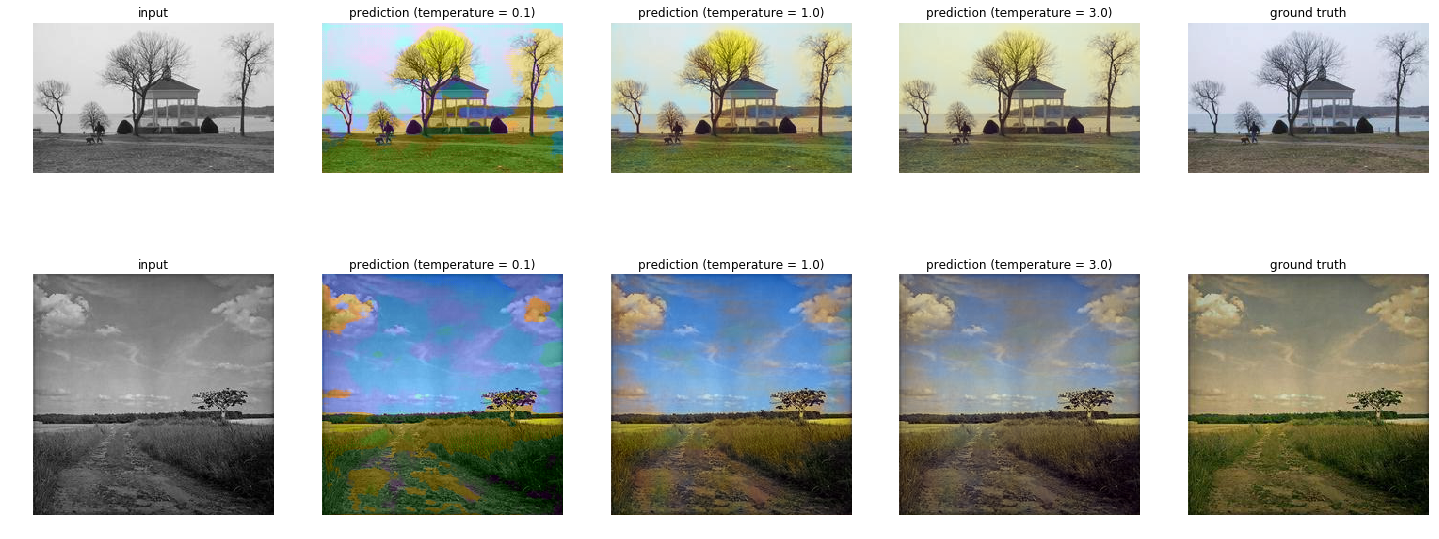}
\caption{Sample predictions of the ColorUNet on the validation set, for bland input images. The ColorUNet's output is more colorful than the ground truth. The bottom example is an old photograph with weared tones.}
\label{better}
\end{center}
\end{figure*}

\section{Dataset and Features}

\subsection{Datasets}

To train our model, we used subsets of the SUN \cite{xiao2010sun} and ImageNet \cite{russakovsky2015imagenet} datasets. We selected 8 categories from ImageNet and 14 categories from SUN, that correspond mainly to nature landscapes. Our final training set is composed of 13,469 images -- as a reference for comparison, \cite{zhang2016colorful} train their model on 1.5M+ images. Our validation set is made of 2,889 images. A sample of the training data is shown in Figure \ref{sampletrain}.
The motivation behing reducing our focus to nature scenes was both conceptual and practical. Conceptual because it seems too optimistic to try to train a rather lightweight model on a very diversified dataset, and practical because it eventually comes down to computing time. We chose to include images representing nature scenes which have pictural elements in common -- such as trees, mountains, clouds, \etc -- as well as color tones -- blue for the sky, sand for beaches, green for grass and forests,  \etc .

We also tried training the model on the full SUN dataset -- roughly 100k images -- (for only one epoch), to compare performance and gain insight on the importance for the model to go over the same examples several times to learn well.

To keep a reasonable size for our tensors and have uniformity in our dataset, the images are all downsampled to fit in a 256$\times$256px frame (if downsampling was necessary). The downsampling is performed using Lanczos method (with a $sinc$ kernel), and we use a mask for non-square images to make sure the loss is relevant.

\subsection{Data Augmentation}

To improve the robustness of the model, we augment our training data with a number of deformations. For each original image of the training set, we generate several training images by combining:
\begin{itemize}
\item Flipping the image along the horizontal axis;
\item Adding noise to the image (with different intensities);
\item Selecting random crops of the image.
\end{itemize}

We expect an improvement of training stability and of the quality of our predictions when including the data augmentation in the pipeline, compared to when simply using more images for training, or training over more epochs. In our case, augmentation dilates the size of our training set sevenfold. We discuss the effect of data augmentation in section \ref{results}.

\section{Results and discussion}\label{results}

\subsection{Evaluation and parameter tuning}

Given the task that we are performing here, there is no relevant metric for the performance of our model. Indeed, the classification loss is useful for training but is not very readable in terms of evaluating the actual prediction performance. Likewise, another loss like the $L_2$ is not relevant, as we discussed in section \ref{relatedwork}. Similarly to \cite{zhang2016colorful}, we rely on human evaluation to evaluate the overall performance of the algorithm, as it is both quick and easy to have a look at a random subsample of the validation set and see if the colorization is satisfying.

The final number of classes chosen for the Color Discretizer is $n=32$, as we want this to be as small as possible (it conditions the size of the output tensor) but 30 color bins are enough to get a very good visual fidelity.

We have chosen the size of our network layers (especially the number of filters) to be of a reasonable order of magnitude, \textit{i.e.} around 32, as our focus was to have a computationally reasonable model. To select the learning rate, we have observed the evolution of the loss on reduced epochs and kept a learning rate that showed a satisfying behavior for the loss. We used an Adam optimizer. Our best model has been trained in 2 steps, with a learning rate decay between the two. Finally, we have chosen a batch size of 64 as it was the largest to fit on the GPUs\footnote{We used a single NVIDIA Tesla K80 GPU} we had: the output tensor has dimensions 256$\times$256$\times$32$\times$\texttt{batch\_size}.

Finally, we used a validation set to evaluate the model. The split between validation and training  was done randomly at preprocessing time.

In addition to visualizing our own colorization results, we visualized the results of the state-of-the-art model, \cite{zhang2016colorful}, which is available online. Unsurprisingly, their results are more realistic. However, on many cases from our validation set, their results are similar to ours. More interestingly, we see that errors are visually similar for the two networks -- \ie color stains -- and that the same kinds of objects cause errors (\eg textures of grass that look like waves on an ocean). This provided a general sanity check for the behavior of our model.
\begin{figure*}
\begin{center}
\includegraphics[width=450px]{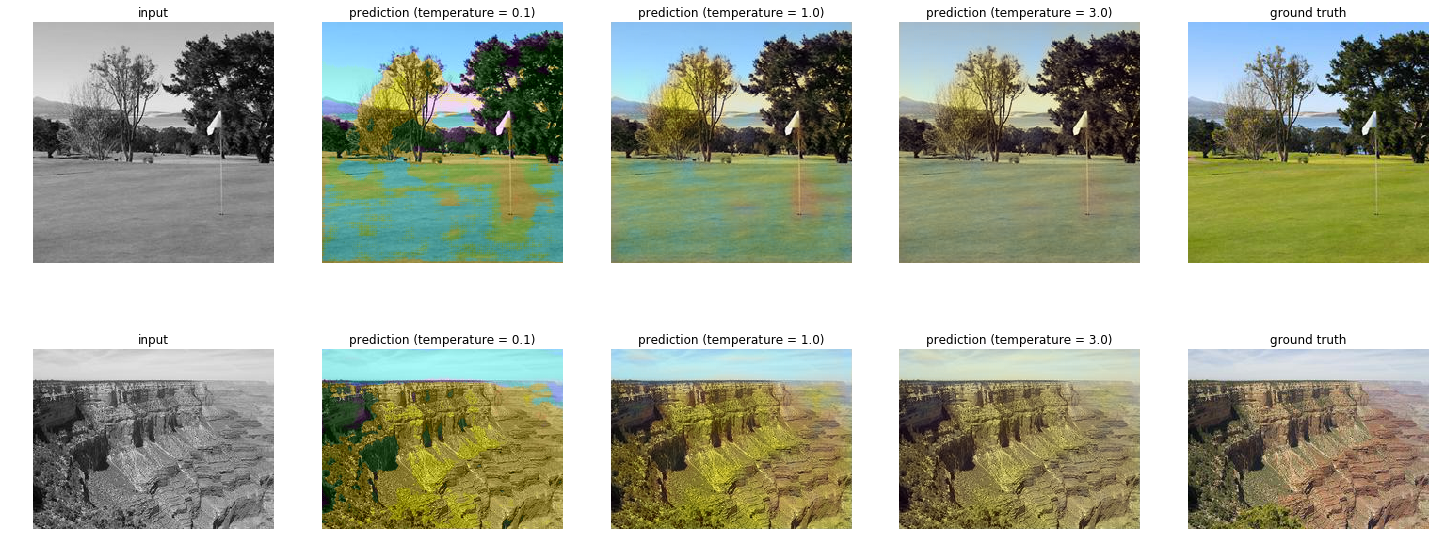}
\caption{Sample bad predictions of the ColorUNet. On the top picture, the network is fooled by the texture of the grass, that has small wavelets similar to sea landscapes, which are vastly represented in the training set. On the bottom picture, the network predicts the canyon as vegetation. It has not learned the color for canyons, both because there are too few in the training set and because the color palette for those landscapes is very specific.}
\label{worse}
\end{center}
\end{figure*}

\begin{figure*}
\begin{center}
\includegraphics[width=450px]{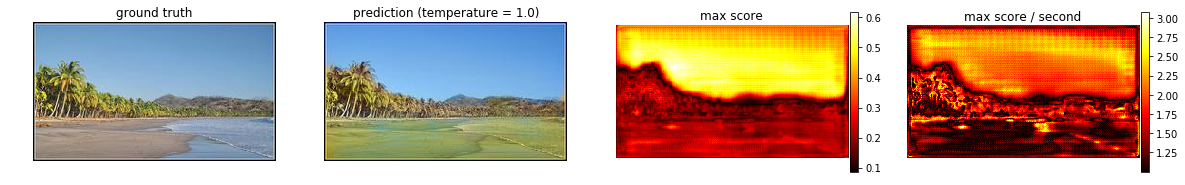}
\caption{Confidence maps for a bad prediction of the ColorUNet. We plot the original image, the prediction with a chosen level of temperature, the value of the top class score, and the ratio between the first and second class scores at each pixel. Dark/red pixels correspond to unsure predictions, whereas yellow/light ones are associated with a high confidence score. The network is fooled by the texture of the sand and sea, that it predicts as grass. On the other hand, the sky is correctly predicted. The confidence maps show that the ColorUNet is sure about its prediction on the sky, but is rather unsure for its prediction of the ground. It is also interesting to note that the regions close to the edges are often more uncertain, although the network performs very well in segmenting parts of the image that should have different colors.}
\label{confidence}
\end{center}
\end{figure*}
\subsection{Colorizing images}
\subsubsection{Overview of the results}

Figure \ref{good} gives some examples of correct colorizations made by the ColorUNet. We can see that the network is able to correctly colorize the main entities of interest -- the sky, trees, grass, water, clouds, \etc --  with vivid colors. The influence of the temperature parameter is very easy to see here, as we have less consistency but more vivid colors for low temperatures (closer to taking the argmax) and smoother, but more desaturated outputs for higher temperatures, as expected in section \ref{methods}.

Interestingly enough, the ColorUNet sometimes gives results that are not similar to the input image, but look better. Those examples are interesting, because they reveal how the coloring model works. Some of them are reported in Figure \ref{better}. This happens with images that have very recognizable inputs, but a limited color palette. In those cases, the ColorUNet's output is more realistic.
Such examples also show that it is difficult to quantatively evaluate the model: indeed, any reasonable computational metric would suggest that these outputs are wrong, while this is clearly the kind of results we are especially interested in.

On the other hand, there are examples where our network is obviously wrong and outputs predictions that are not plausible. Some such examples are seen in \ref{worse}.

Overall, our ColorUNet performs well at the limited task that we want to tackle. Not very surprisingly, it does not perform well on new categories, and especially on indoor scenes.

\subsubsection{Further analysis}

We propose some visualizations to further analyze the behavior of the model.

\paragraph{Confidence maps}

In order to better understand the behavior of our model, we have designed a \textit{confidence map}, showed in Figure \ref{confidence}. We use the value of the top bin's score and its relative magnitude compared to the second score of each pixel to visualize the confidence of the prediction. This is also a good sanity check. Actually, given that our model is intended to be lightweight, we do not expect it to make outstanding predictions on all the pictures. However, we expect our model to be unsure about the wrong results and the difficult parts of the images.

\paragraph{Loss analysis}

\begin{figure}
\begin{center}
\includegraphics[width=220px]{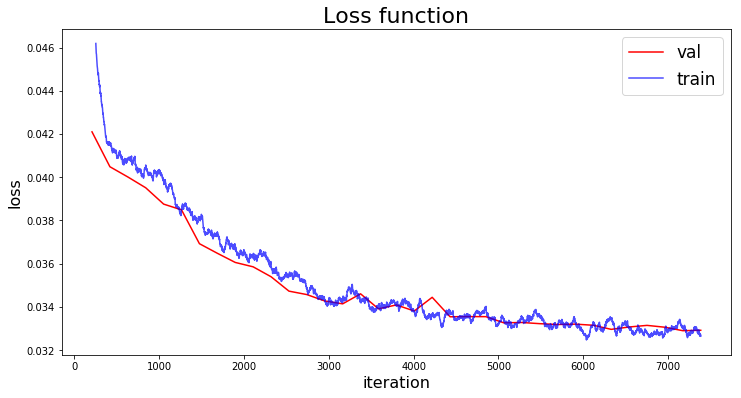}
\caption{Evolution of the loss of ColorUNet accross the learning iterations. The validation (red) loss matches the train (blue) loss, which shows that we have not overfitted the model to the training set.}
\label{lossplot}
\end{center}
\end{figure}

The ability of the model to generalize can be explored from a quantitative point of view. Figure \ref{lossplot} shows that the performance of our model on the validation set matches that obtained on the training set, in terms of loss. This result proves that we avoid overfitting and validates our initial idea of aiming for a lightweight model, which cannot be over-parametric. This is also promising since it implies that there is room for progress. The ColorUNet can be further trained, and kernel sizes in intermediate layers can be increased, which will likely yield even better results with more computing power.

When taking a closer look at in-sample examples, we see that overall the quality of the prediction is slightly better but comparable to what we have in the validation set. This is extra evidence that we have not overfitted to the training set. However, given the evolution of the loss, we can even expect our model to be slightly underfitted, and it may benefit from further training. However, given the limited time we had to complete this project, we preferred focusing on other experiments (such as studying the effect of data augmentation or training on an extended dataset) rather than spending time overfitting the model.

\paragraph{Color distribution analysis}

\begin{figure}
\begin{center}
\includegraphics[width=250px]{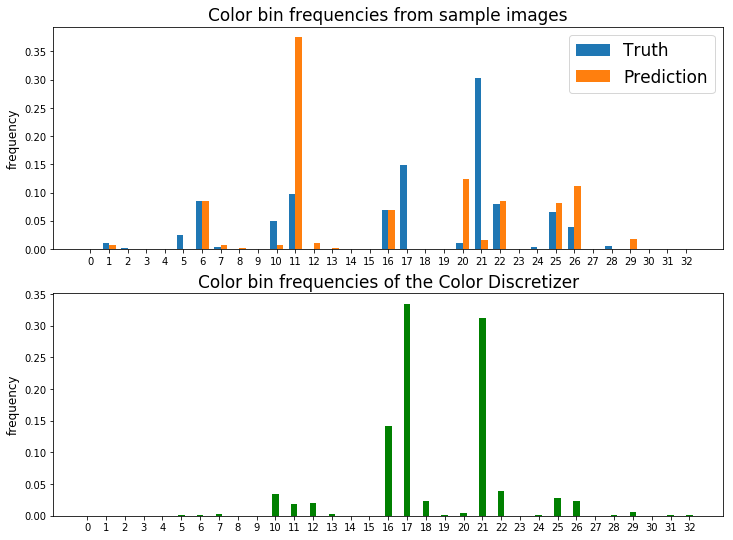}
\caption{Histograms of the frequencies of the 32 selected color bins. Top: frequencies in the pixels of 500 sample images; true color channels (blue); colors predicted by ColorUNet \text{(orange)}. \text{Bottom}: frequencies of the color bins in the Color Discretizer (green), representing the whole dataset. The ColorUNet differs from the ground truth which has ``common'' colors (bins 17, 21), it prefers ``rare'' colors (bins 11, 20, 26). }
\label{histogram}
\end{center}
\end{figure}

In order to understand the impact of our decision to encourage the prediction of ``rare'' colors by weighting the loss, we produced histograms of the frequencies of the 32 color bins (see Figure \ref{histogram}). We see that the prediction of the ColorUNet does not exactly match the ground truth. By comparing the histograms with the frequencies in the whole dataset, we see that the ground truth contained ``common'' colors. The ColorUNet chooses to aim for ``rare'' colors, thus producing vivid images. Since this choice does not impair the realism of outputs, we believe that our contribution achieved our goal of colorizing with lively colors.

\paragraph{Influence of data augmentation}

To evaluate the influence of data augmentation, we have run 2 models on the same training set, with and without data augmentation -- increasing the number of epochs sevenfold for the latter, so that both models train over the same number of iterations. As expected, data augmentation brings more stability to the predictions, especially for the low temperatures.

\begin{figure}
\begin{center}
\includegraphics[width=220px]{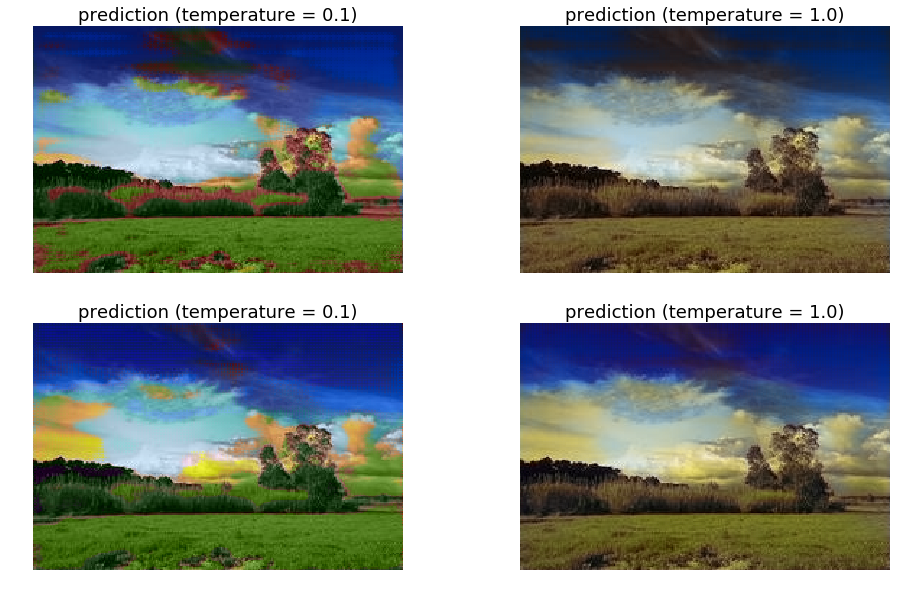}
\caption{Influence of data augmentation. Top: prediction for 2 levels of temperature, with data augmentation. Bottom:  prediction for 2 levels of temperature, without data augmentation. The data augmentated model tends to predict finer and smaller color stains that are closer to the actual coloring, which is easy to see on the low temperature prediction on the left. This results in an overall enhancement of the final prediction (on the right), where the sky tones are more realistic}
\label{dacomp}
\end{center}
\end{figure}

Figure \ref{dacomp} illustrate the result of this experiment.

\subsection{Application to video colorization}

The last task we tried to tackle is colorizing greyscale videos. The natural way to tackle this problem is to extract individual frames that constitute the video, and to colorize each frame with the previous ColorUNet. Converting this series of colorized images back into a video produces a colorized video. The results of this work can be seen on our \href{https://github.com/vincentbillaut/all-colors-matter}{Github repo's homepage}.

One understands here that this preliminary work does not incorporate the structure of the video. However, this structural information might be crucial for two reasons. First, as the movie evolves over a progressively moving scene, we want some stability over the set of colors that are predicted. From one frame to the next, the model should colorize in a similar way the corresponding items. Second, an algorithm having an extra temporal information on top of the single-frame prediction might even perform better. Indeed, including past information enables a better understanding of the context of the image, and the incorporation of content that is now hidden from the frame.

Consequently, the method we will use should know what was predicted before and take this knowledge into account when deciding for the next frame's color. We choose both to stabilize predictions and to incorporate past information, and to that effect smooth the predictions by applying a temporal kernel, which weighs the past predictions' probabilities with an exponential decay:

 $$\hat p_t = \sum_{i = 0}^T p_{t-i} e^{-\alpha i}.$$

The 32-class probability vector we use for the video colorization, $\hat p_t$, aggregates past frames' probabilities $p_{t-i}$, over $T=20$ frames, with a decay coefficient of $\alpha=0.2$. The results are smoother and somehow more natural.

\subsection{Extending the training set}

As a sanity check, we have also trained a model using the full SUN dataset, which is 10 times bigger than our data-augmented training set. Because of time constraints, we have trained it on fewer epochs. The training loss was similar. However, we note that the results were very poor, and that the network tended to output very desaturated images. Actually, compared to the baseline model of \cite{zhang2016colorful}, our network structure has few filters in its deepest layers. It is therefore unable to capture as many features as needed to be able to recognize the variety of scenes present in the full dataset. This confirms that our architecture is a lightweight version, that is adapted to a reduced variety of scenes, and that trains rapidly. However, we could expect that applying the same structure with more filters, to have a complexity closer to \cite{zhang2016colorful} could allow to have a much quicker training on a full dataset.

An interesting next step to take would be to do a full benchmark of the model performance over different ranges of variety in the training set. Our lightweight model does not perform well on many classes because it is not complex enough. However, too few classes do not allow to state an interesting problem: \eg learning on a set of beach pictures can be done with bimodal predictions -- blue or yellow. An extensive quantitative benchmark of this effect could be interesting but very time and resource-consuming, and beyond the scope of this project.
\section{Conclusion and perspectives}

In this work, we formulated the colorization problem as a classification problem to get colorful results. The ColorUNet we have designed is a lightweight version of the state-of-the-art architecture that allows to quickly get satisfying results on a reduced dataset, with limited computational power.

Our predictions have similar limitations as the reference model. Extending the task to more categories would require more parameters, and our experiments have shown that a U-Net inspired architecture enhances and accelerates the learning process for the colorization problem. We propose a lightweight extension to colorize video sequences.

Possible extensions would include training a network with more filters on more classes, but it would most likely yield similar results to \cite{zhang2016colorful}, and still have spatial consistency problems. Using the confidence maps we designed to automatically select seed pixels and apply the heuristics described in \cite{levin2004colorization} to \textit{diffuse} those seeds would be an elegant way to post-process our images and have realistic, consistent but yet colorful results, taking the best of both ``traditional'' and ``convolutional'' approaches to colorization.

Combining old optimization methods and convolutional colorization could also be a way to tackle video colorization, by adapting the temporal consistency heuristics of \cite{zhu2017video} to the \textit{seed pixels} generated by our network. Enforcing temporal consistency of the seeds used to postprocess the recolorized image would be a proxy to enforce temporal consistency for the colorization overall.

\section*{Contributions and acknowledgements}

We would like to thank the CS231n teaching staff for giving us access to the Google Cloud Compute platform and for the teaching material.

This project was implemented using \texttt{python}, and notably the  TensorFlow library (\cite{abadi2016tensorflow}).

All authors contributed equally to this work.

{\small
\bibliographystyle{ieee}
\bibliography{references}
}

\end{document}